\def\year{2020}\relax
\documentclass[letterpaper]{article} % DO NOT CHANGE THIS
\usepackage{aaai20}  % DO NOT CHANGE THIS
\usepackage{times}  % DO NOT CHANGE THIS
\usepackage{helvet} % DO NOT CHANGE THIS
\usepackage{courier}  % DO NOT CHANGE THIS
\usepackage[hyphens]{url}  % DO NOT CHANGE THIS
\usepackage{graphicx} % DO NOT CHANGE THIS
\usepackage{graphicx}  % DO NOT CHANGE THIS
\usepackage{makecell}

\title{Attention-based Multi-modal Fusion Network for Semantic Scene Completion}
\author{
Siqi Li\textsuperscript{\rm 1}, 
Changqing Zou\textsuperscript{\rm 2}, 
Yipeng Li\textsuperscript{\rm 3}, 
Xibin Zhao\textsuperscript{\rm 1}, 
Yue Gao\textsuperscript{\rm 1}\thanks{Corresponding author} 
\\ \textsuperscript{\rm 1}BNRist, KLISS, School of Software, Tsinghua University, China 
\\ \textsuperscript{\rm 2}Sun Yat-sen University, \textsuperscript{\rm 3}Department of Automation, Tsinghua University, China 
\\ lsq19@mails.tsinghua.edu.cn, aaronzou1125@gmail.com, \{liep, zxb, gaoyue\}@tsinghua.edu.cn
}

\begin{document}

\maketitle

\begin{abstract}
This paper presents an end-to-end 3D convolutional network named attention-based multi-modal fusion network (AMFNet) for the semantic scene completion (SSC) task of inferring the occupancy and semantic labels of a volumetric 3D scene from single-view RGB-D images. Compared with previous methods which use only the semantic features extracted from RGB-D images, the proposed AMFNet learns to perform effective 3D scene completion and semantic segmentation simultaneously via leveraging the experience of inferring 2D semantic segmentation from RGB-D images as well as the reliable depth cues in spatial dimension. It is achieved by employing a multi-modal fusion architecture boosted from 2D semantic segmentation and a 3D semantic completion network empowered by residual attention blocks. We validate our method on both the synthetic SUNCG-RGBD dataset and the real NYUv2 dataset and the results show that our method respectively achieves the gains of $2.5\%$ and $2.6\%$ on the synthetic SUNCG-RGBD dataset and the real NYUv2 dataset against the state-of-the-art method.

\end{abstract}

\section{Introduction}
Understanding and reconstructing a 3D scene from partial observations is a very important technique which has received increasing research attention in recent years due to its commercial potential in a large variety of robotics and vision tasks such as robotic navigation~\cite{Gupta2013PerceptualOA}, autonomous driving~\cite{LaugierPPYYTMN11}, and scene reconstruction~\cite{Hays2008SceneCU,han2019deep}. Given a single depth image or RGB-D images of a 3D scene, many papers~\cite{Gupta2013PerceptualOA,Ren2012RGBDSL,Firman2016StructuredPO} have been proposed to complete or segment the 3D scene with neural networks. More recently, a set of methods~\cite{Song2016SemanticSC,Liu2018SeeAT,Tong2018,Zhang2018EfficientSS,Li2019RGBDBD} have been developed to automatically predict the semantic labels, together with completing the 3D geometry, of the objects in a 3D scene from a single view of the scene using convolutional neural networks.

%=====================Figure=====================
\begin{figure}[t]
\centering
\includegraphics[width=0.98\columnwidth]{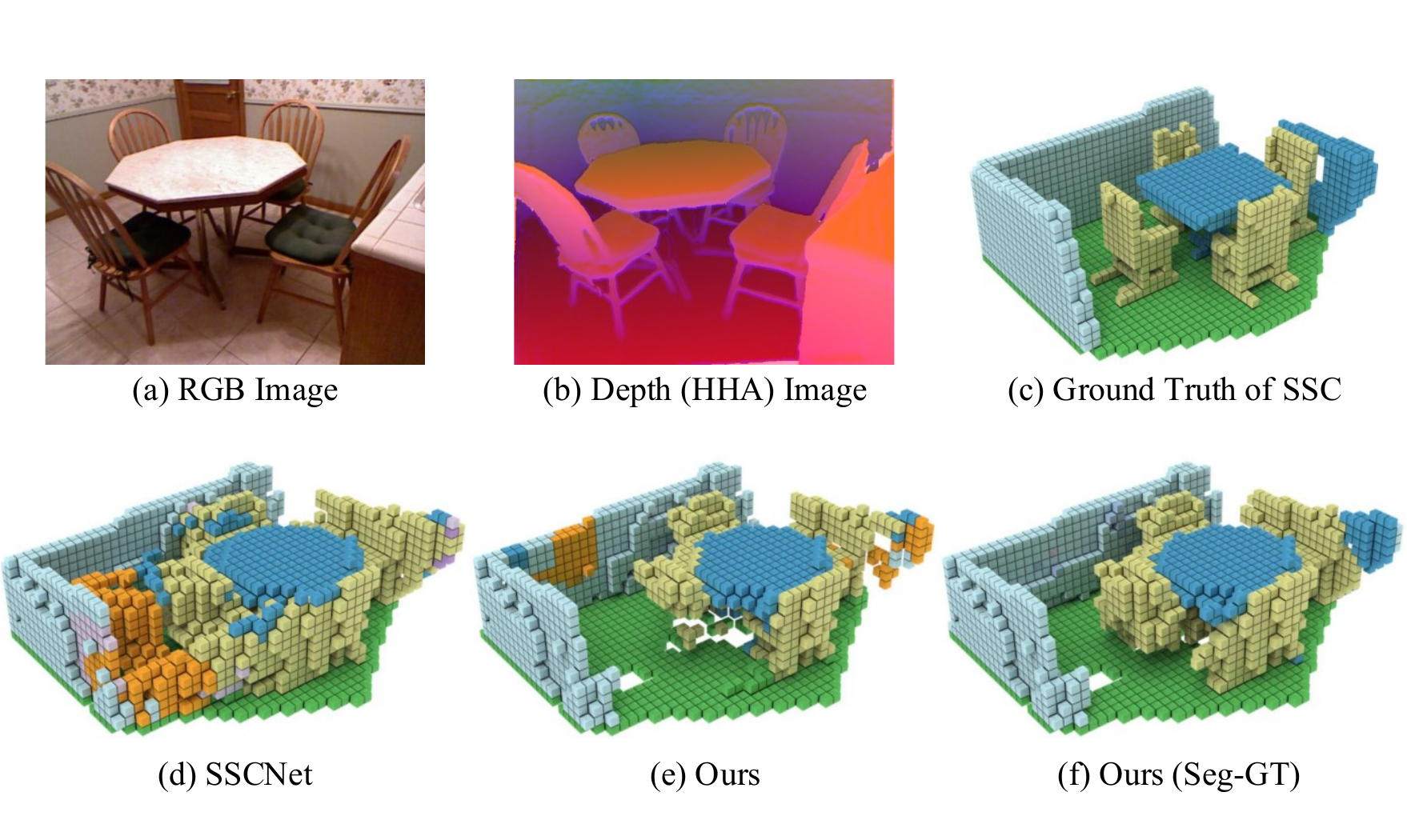}
\caption{Given RGB-D images (a-b) as input, the proposed AMFNet can produce more accurate scene completion and scene segmentation result (e) than the previous methods (e.g., SSCNet~\cite{Song2016SemanticSC} (d)). Directly boosting the AMFNet from the 2D segmentation ground truth of the input RGB-D images can even produce a result closer to the ground truth (c). }
\label{fig1}
\end{figure}

\begin{figure*}[t]
\centering
\includegraphics[width=0.95\textwidth]{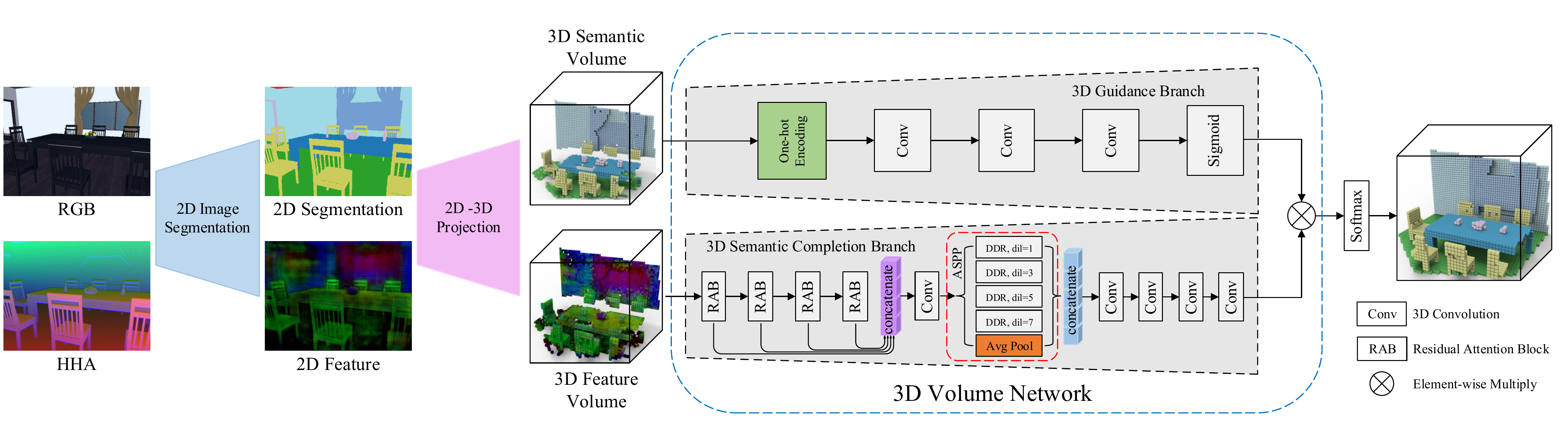}
\caption{Architecture of AMFNet. Taking RGB-D images (separated to a RGB and a HHA image) as input, {AMFNet predicts voxel occupancy and object labels of the scene simultaneously.} It boosts the 3D completion and segmentation from an initial 3D semantic feature volume produced by computing the 2D-3D projection of the results of a 2D segmentation network. } 
\label{network_structure}
\end{figure*}
%=====================Figure=====================

This line of literature focusing on the task of semantic scene completion (SSC) was initiated by those methods~\cite{Song2016SemanticSC,Zhang2018EfficientSS} which take as input only the depth information. Several later works~\cite{Liu2018SeeAT,Li2019RGBDBD,Hilton2018} argue that the RGB color information, which is often captured together with the depth information by an RGB-D sensor, could be used to improve the performance of the networks which take only the depth information. The experimental results from these works reveal that the RGB color channels provide complementary information to the depth information and higher accuracy of SSC can be achieved by using the fused information.

In this work, we propose a novel approach for the task of semantic scene completion, which makes use of not only the information of RGB color channels and depth but also the “experience” of deep network learning 2D semantic segmentation. Specifically, we train a model that leverages 2D semantic segmentation information to guide both 3D completion and semantic labeling (semantic segmentation) in the SSC task. Some prior works~\cite{Chen2018EncoderDecoderWA,yang2018denseaspp,li2019dfanet,xiong2019upsnet} show that 2D segmentation is more accurate than the 3D SSC task. Our model is based on the assumption that the image features which are reliable for 2D semantic segmentation should also be reliable for semantic scene completion. With this assumption, we propose a novel two-branch multi-modal fusion network where the 3D segmentation information, which is boosted from 2D segmentation, could guide both 3D completion and semantic labeling feasibly. Moreover, due to the inherent sparsity of 3D data (most voxels in a 3D scene are empty), we believe that some features (e.g., color and texture) associated with the empty voxels are valueless. Therefore, we propose a residual attention block (RAB) based 3D network to make full use of reliable depth cues in the spatial dimension for semantic scene completion.

The proposed network contains three sequential modules: a 2D segmentation module which extracts 2D image features and produces 2D semantic labels, a 2D-3D projection layer which generates a 3D semantic feature volume from the output of the 2D segmentation module, and an attention-driven two-branch 3D volume network (named 3D volume network) which infers a complete 3D volume with semantic labels from the initial 3D scene.

We validate our approach on both the synthetic SUNCG-RGBD dataset~\cite{Liu2018SeeAT} and the real NYUv2 dataset~\cite{Silberman2012IndoorSA}, and the results show that our method achieves a gain of 2.5\% on the SUNCG-RGBD dataset and a gain of 2.6\% on the NYUv2 dataset against the state-of-the-art method~\cite{Li2019RGBDBD}. Our analytical study also observes that directly boosting the AMFNet from the 2D segmentation ground truth of the input RGB-D images (also see Figure~\ref{fig1}(f)) can achieve a gain of 3.8\% on the synthetic SUNCG-RGBD dataset and a gain of 4.3\% on the real NYUv2 dataset against the state-of-the-art method~\cite{Li2019RGBDBD}. 

Our main contributions can be summarized as follows:
\begin{itemize}
\item A multi-modal fusion network boosted from 2D semantic segmentation is proposed for semantic scene completion. Compared to previous works~\cite{Song2016SemanticSC,Liu2018SeeAT,Li2019RGBDBD}, the major advantage of this network is that 2D semantic segmentation is used to guide and improve the 3D feature extraction for the SSC task. We demonstrate this advantage can significantly boost the overall performance of the task of SSC by improving both scene completion and segmentation. 

\item A residual attention block (RAB) based network is proposed for scene semantic completion. Our experiments show that the proposed RAB is particularly effective for some particular objects in a 3D scene.

\item The proposed end-to-end framework achieves state-of-the-art performance on two large-scale datasets, i.e., the NYUv2~\cite{Silberman2012IndoorSA} and the SUNCG-RGBD~\cite{Liu2018SeeAT} datasets.
\end{itemize}

\section{Related Works}
\paragraph{Semantic Scene Completion} 
The pioneering work for the task of semantic scene completion was proposed by Song et al. in \cite{Song2016SemanticSC}. Although this method achieves good performance in terms of performing both scene completion and semantic segmentation simultaneously from a single-view depth image within a task, it requires an extremely time-consuming data pre-processing step of computing the flipped truncated signed distance function (fTSDF) used for eliminating intense gradients and view dependency.

Motivated by the fact that most of the voxels in the 3D volume are empty and useless,
Zhang et al. further improved the SSCNet by applying a spatial group convolution (SGC) in EsscNet~\cite{Zhang2018EfficientSS}. This method divides the input fTSDF into different groups and then forwards them to a 3D sub-manifold sparse convolutional network \cite{Graham20173DSS}. A limitation of the EsscNet is that the input features are empirically divided into several groups, which might lead to performance degradation.

More close to our work, SATNet~\cite{Liu2018SeeAT} is a model that performs the SSC task from 2D semantic features. This model contains three sequential modules similar to the proposed AMFNet: a 2D network to extract semantic features from RGB-D images, a 2D-3D projection layer, and a 3D network to complete the scene. Generally, the SATNet has a data processing flow (i.e., boosting 3D completion and segmentation from the 2D-3D projection of the 2D semantic features) similar to the 3D-semantic completion branch (the bottom one in Figure~\ref{network_structure}) of AMFNet. DDRNet~\cite{Li2019RGBDBD} is a light-weight deep model with an architecture similar to SATNet, of which the major contribution is a dimensional decomposition residual (DDR) block, which reduces the network parameters dramatically by decomposing the traditional 3D convolution block into consecutive layers channel-wise. The proposed AMFNet follows data processing flow similar to SATNet and DDRNet, but rather than perform 3D scene completion and semantic segmentation in a single branch, AMFNet employs two separated branches for 3D scene completion and semantic segmentation respectively, which enables the 2D semantic segmentation information to be used to explicitly guide the 3D scene completion and semantic segmentation simultaneously.

\paragraph{Attention Mechanism}
A large number of CNN-based methods~\cite{Wang2017ResidualAN,Hu2017SqueezeandExcitationN} use the attention mechanism to improve the performance of the classification task. Hu et al.~\cite{Hu2017SqueezeandExcitationN} employ an attention scheme which infers the importance of multiple channels for image classification. This method achieves significant gains in classification accuracy by making the model focus on meaningful regions of the image and ignores meaningless regions with the attention scheme. Woo et al.~\cite{Woo2018CBAMCB} propose an attention scheme named convolutional bottleneck attention module (CBAM) to estimate channel-wise attention weight and spatial-wise attention weight. Inspired by the great success of the application of attention mechanism in the image classification task and the inherent sparsity of 3D data, we believe that the ability of the attention mechanism, which makes the model focus on important parts, maybe also helpful for the task of SSC. In our network, to inject the attention into the 3D volume network without significant parameter increase, we graft the CBAM attention module onto a computation-efficient residual block to form a novel attention-injected residual block (RAB) and use this RAB for 3D feature extraction. 

%=====================Figure=====================
\begin{figure}[t]
\centering
\includegraphics[width=0.99\columnwidth]{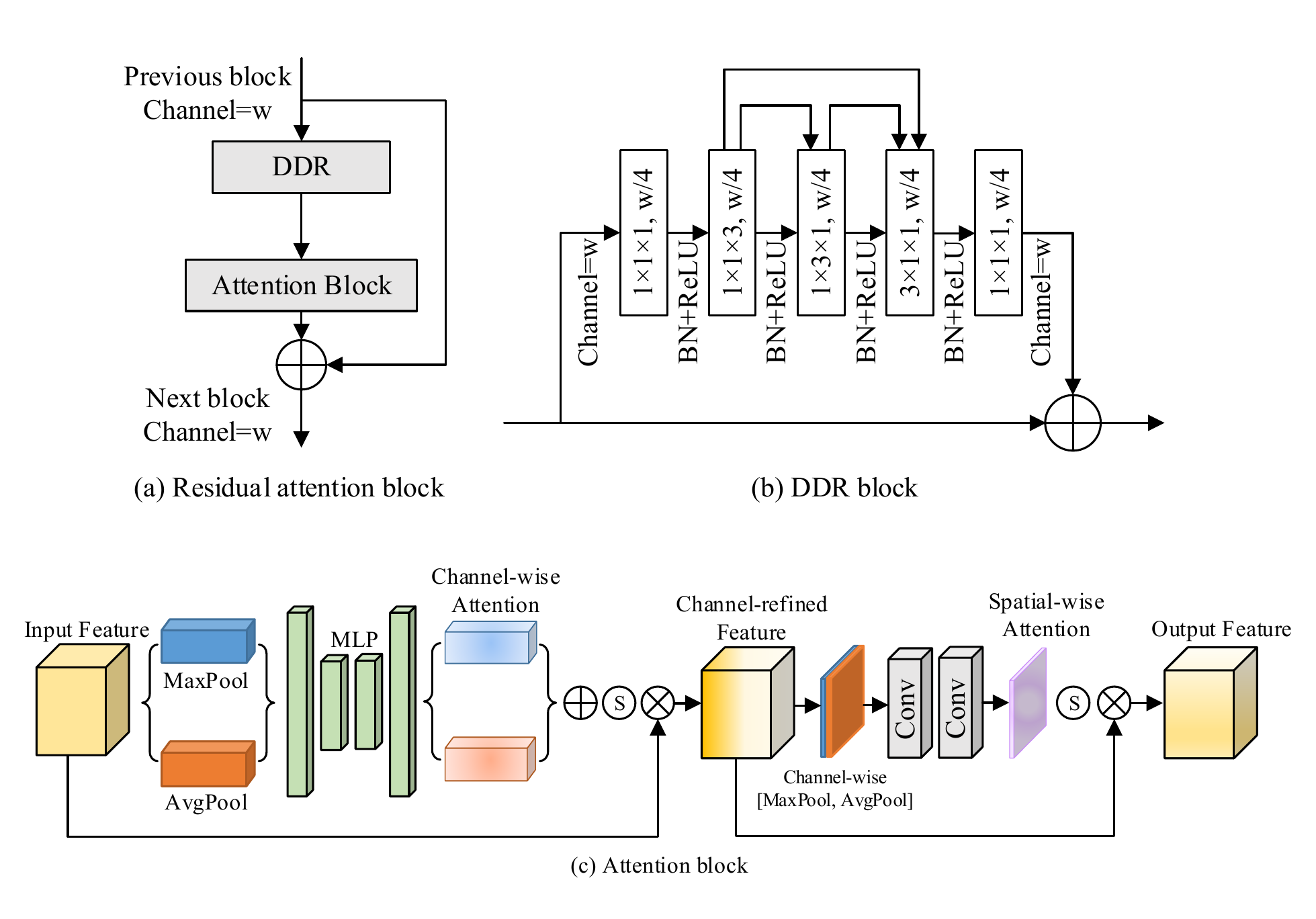}
\caption{Illustration of the proposed residual attention block (RAB). RAB has a structure similar to the DDR block~\cite{Li2019RGBDBD} but with both channel-wise and spatial-wise attention injected. 
}
\label{RAB_visualization}
\end{figure}
%=====================Figure=====================

\section{Method}
In this section, we first present the structure of the proposed AMFNet and then detail each module. 
 
The proposed AMFNet, as illustrated in Figure~\ref{network_structure}, mainly contains three sequential modules: a 2D segmentation network, a 2D-3D projection layer, and a two-branch 3D volume network. This network takes single-view RGB-D images as input and outputs occupancy and semantic labels for all voxels in the scene. The whole network can be trained in an end-to-end manner. We next introduce each module in the sequence of the flow of data processing.

\subsection{2D Segmentation Network}
\label{2D Network}
The 2D segmentation network extracts 2D geometry features and performs 2D semantic segmentation from the input RGB-D images. Since it is unlikely that a CNN would automatically learn to compute the properties of geocentric pose that emphasize complementary discontinuities in the RGB-D image (e.g., depth, surface normal and height), we first compute a three-channel HHA \cite{Gupta2014LearningRF} encoding result from the input depth image. After that, the RGB image and HHA image are fed into two 2D segmentation networks that have the same structure but do not share weights, and their individual outputs are fused together by a convolution layer with a kernel size of 1. In our implementation, the 2D segmentation network has an encoder-decoder structure where the encoder uses ResNet-101~\cite{He2015DeepRL}, and the decoder contains a series of up-sampling convolutions and an atrous spatial pyramid pooling (ASPP) module~\cite{Chen2016DeepLabSI,Chen2018EncoderDecoderWA} which can extract multi-scale features. The output of the 2D segmentation network is a feature map with 64 channels and a semantic segmentation map.

\subsection{2D-3D {Projection} Layer}
\label{projection layer}
We employ a projection layer similar to those in SATNet~\cite{Liu2018SeeAT} and DDRNet~\cite{Li2019RGBDBD} to generate an initial 3D semantic feature volume (can be separated into a 3D semantic volume and a 3D feature volume, as shown in Figure~\ref{network_structure}). For each input RGB-D image, given the intrinsic and extrinsic camera matrix, a 3D volume can be directly computed from the depth information (see SATNet~\cite{Liu2018SeeAT} for reference). Since each pixel in the RGB-D images correspond to a tensor in the 2D feature map as well as a semantic label in the 2D segmentation map, every feature tensor and semantic label can be projected into the 3D volume at the location with the same depth value. In this way, we obtain a view-independent 3D semantic feature volume and a 3D semantic volume (only visible voxels have semantic labels), as shown in Figure~\ref{network_structure}. The 2D-3D projection operation can be recognized as reshaping the feature maps during the forward propagation. During back propagation, we only map the gradient of the voxels on the visible surface (discarding the gradient of other voxels) to the 2D pixels and continue the back propagation in the 2D network.

\subsection{3D Volume Network} 
The 3D volume network, which takes the output of the 2D-3D projection layer as input, contains two branches: one for 3D guidance information and the other for 3D semantic completion. At the end of the two branches, the outputs of the guidance branch and semantic completion branch are element-wise multiplied and forwarded into a softmax layer to generate a 3D volume with semantic labels.

\subsubsection{3D-Guidance Branch}
\label{3DSegm} 
The 3D-guidance branch is used to provide the guidance information for the branch of 3D semantic completion, which is boosted from an initial 3D semantic volumetric scene where visible voxels have initial semantic labels.
The initial 3D semantic volume is first encoded by a one-hot encoder to achieve an ROI region (3D bounding box) for a specific category from the initial 3D semantic volume. For a dataset containing $N$ different categories, the 3D semantic volume could be encoded by a volume with $N$ channels. Each channel of the volume can be used to represent the ROI region of a specific category. For each channel, the value of the voxels within the bounding box of the corresponding category is set to one, otherwise, it is set to zero. The one-hot encoding introduces spatial boundary constraints into the network for each category, which improves the prediction of 3D semantic volume. (See the experimental analysis for more details.) The encoded volume is then fed into three 3D dense convolution layers and a sigmoid layer to obtain the output features of the 3D-guidance branch. It is worth mentioning that we do not perform the one-hot encoding with instance-level semantic information which may provide more accurate boundary constraints because both datasets in our experiment do not provide instance-level segmentation information. 

%=====================Table=====================
\begin{table*}
\centering
\resizebox{0.99\textwidth}!{
\begin{tabular}{ l | ccc |cccccccccccc } 
\Xhline{1.2pt}
 & \multicolumn{3}{c|}{scene completion} & \multicolumn{12}{c}{semantic scene completion} \\ \hline
 Method & prec. & recall & IoU & ceil. & floor & wall & win. & chair & bed & sofa & table & tvs & furn. & objs. & avg. \\ \hline
\begin{tabular}[c]{@{}c@{}}Lin et al. \shortcite{Lin2013HolisticSU}\end{tabular} & 58.5 & 49.9 & 36.4 & 0.0 & 11.7 & 13.3 & 14.1 & 9.4 & 29.0 & 24.0 & 6.0 & 7.0 & 16.2 & 1.1 & 12.0 \\
\begin{tabular}[c]{@{}c@{}}Geiger et al. \shortcite{Geiger2015Joint3O}\end{tabular} & 65.7 & 58.0 & 44.4 & 10.2 & 62.5 & 19.1 & 5.8 & 8.5 & 40.6 & 27.7 & 7.0 & 6.0 & 22.6 & 5.9 & 19.6 \\
\begin{tabular}[c]{@{}c@{}}SSCNet \shortcite{Song2016SemanticSC}\end{tabular} & 57.0 & \textbf{94.5} & 55.1 & 15.1 & \textbf{94.7} & 24.4 & 0.0 & 12.6 & 32.1 & 35.0 & 13.0 & 7.8 & 27.1 & 10.1 & 24.7 \\
\begin{tabular}[c]{@{}c@{}}EsscNet \shortcite{Zhang2018EfficientSS}\end{tabular} & \textbf{71.9} & 71.9 & 56.2 & 17.5 & 75.4 & 25.8 & 6.7 & 15.3 & 53.8 & 42.4 & 11.2 & 0 & 33.4 & 11.8 & 26.7 \\
\begin{tabular}[c]{@{}c@{}}DDRNet \shortcite{Li2019RGBDBD}\end{tabular} & 71.5 & 80.8 & \textbf{61.0} & 21.1 & 92.2 & \textbf{33.5} & 6.8 & 14.8 & 48.3 & 42.3 & 13.2 & \textbf{13.9} & 35.3 & 13.2 & 30.4 \\ \hline
Ours & 67.9 & 82.3 & 59.0 & 16.7 & 89.2 & 27.3 & \textbf{19.2} & \textbf{20.2} & \textbf{56.1} & \textbf{50.4} & \textbf{15.1} & 13.5 & \textbf{36.8} & 18.0 & \textbf{33.0} \\
\begin{tabular}[c]{@{}c@{}}Ours (w/o-Attn)\end{tabular} & 64.5 & 86.5 & 58.6 & \textbf{21.3} & 90.3 & 26.1 & 7.7 & 18.0 & 53.8 & 48.4 & 13.0 & 0 & 36.7 & 16.3 & 30.1 \\
\begin{tabular}[c]{@{}c@{}}Ours (w/o-Seg)\end{tabular} & 68.5 & 78.6 & 57.3 & 13.7 & \textbf{94.7} & 26.9 & 14.8 & 12.7 & 48.5 & 39.8 & 10.4 & 12.2 & 36.2 & \textbf{19.2} & 29.9 \\
\begin{tabular}[c]{@{}c@{}}Ours (basic)\end{tabular} & 64.0 & 86.6 & 58.0 & 20.9 & 94.1 & 27.1 & 15.7 & 12.4 & 46.0 & 45.1 & 13.9 & 0.3 & 32.3 & 15.7 & 29.4 \\ \hline
\begin{tabular}[c]{@{}c@{}}Ours (Seg-GT)\end{tabular} & 66.3 & 80.5 & 57.2 & 20.0 & 78.7 & 27.3 & 20.5 & 21.8 & 56.5 & 53.9 & 19.5 & 18.8 & 40.1 & 19.5 & 34.2 \\ 
\Xhline{1.2pt}
\end{tabular}}
\caption{Results on the NYUv2 dataset. Bold numbers represent the best scores.}
\label{tab1} 
\end{table*}

\begin{table*}
\centering
\resizebox{0.99\textwidth}!{
\begin{tabular}{ p{2.75cm} |ccc|cccccccccccc } 
\Xhline{1.2pt}
& \multicolumn{3}{c|}{scene completion} & \multicolumn{12}{c}{semantic scene completion} \\ \hline
Method & prec. & recall & IoU & ceil. & floor & wall & win. & chair & bed & sofa & table & tvs & furn. & objs. & avg. \\ \hline
SSCNet \shortcite{Song2016SemanticSC} & 43.5 & 90.7 & 41.5 & 64.9 & 60.1 & \textbf{57.6} & 25.2 & 25.5 & 40.4 & 37.9 & 23.1 & 29.8 & \textbf{45.7} & 4.7 & 37.7 \\
SATNet \shortcite{Liu2018SeeAT}& 56.7 & 91.7 & 53.9 & 65.5 & 60.7 & 50.3 & 56.4 & 26.1 & \textbf{47.3} & 43.7 & 30.6 & 37.2 & 44.9 & \textbf{30.0} & 44.8 \\ \hline
Ours & \textbf{57.5} & 91.6 & \textbf{54.5} & \textbf{80.4} & \textbf{69.1} & 55.0 & \textbf{60.4} & \textbf{27.0} & 42.2 & 46.7 & \textbf{32.5} & 42.3 & 36.9 & 27.4 & \textbf{47.3} \\
\begin{tabular}[c]{@{}c@{}}Ours (w/o-Attn)\end{tabular} & 54.8 & 94.7 & 53.2 & 79.6 & 66.9 & 51.7 & 60.2 & 26.5 & 38.2 & 45.5 & 24.5 & 27.9 & 38.1 & 28.7 & 44.3 \\
\begin{tabular}[c]{@{}c@{}}Ours (w/o-Seg)\end{tabular} & 54.1 & 94.0 & 52.4 & 76.6 & 61.0 & 53.7 & 54.7 & 26.2 & 37.3 & \textbf{49.4} & 31.8 & 43.7 & 40.7 & 25.0 & 45.5\\
\begin{tabular}[c]{@{}c@{}}Ours (basic)\end{tabular} & 47.5 & \textbf{96.4} & 46.7 & 71.0 & 49.7 & 41.4 & 57.3 & 21.6 & 41.3 & 46.1 & 30.1 & \textbf{44.8} & 39.9 & 21.4 & 42.2\\ \hline
\begin{tabular}[c]{@{}c@{}}Ours (Seg-GT)\end{tabular} & 60.6 & 89.1 & 56.3 & 81.3 & 68.5 & 54.1 & 61.8 & 30.2 & 45.9 & 50.7 & 34.3 & 42.7 & 41.9 & 28.4 & 49.1 \\
\Xhline{1.2pt}
\end{tabular}}
\caption{Results on the SUNCG-RGBD dataset (we do not compare our method to EsscNet and DDRNet because neither of them report their performances on this dataset). Bold numbers represent the best scores.} 
\label{tab2}
\end{table*}
%=====================Table=====================

\subsubsection{3D Semantic Completion Branch}
The 3D-semantic completion branch, which takes the initial 3D feature volume as input, is mainly used to infer the voxel occupancy of the 3D scene. The data processing flow is illustrated in Figure~\ref{network_structure}. 
The input 3D features are firstly forwarded to four RAB modules, and the outputs of each block are concatenated and forwarded to a 3D convolution layer with a kernel size of $1$ for multi-level feature fusion. After that, the fused feature is further forwarded to a 3D ASPP block (i.e., multiple parallel atrous convolutional layers with different dilatation rates, which is a powerful way to handle objects with various sizes) to exploit multi-scale features. The output of the ASPP block is lastly forwarded to four cascaded 3D convolution layers to generate high-level features for the fusion with the 3D-Guidance branch. 
In this branch, the attention focusing on reliable depth cues in the spatial dimension are mainly captured by the RAB modules. Our experiments in the next section will show the attention scheme contributes much to the overall performance ($3\%$ gains on average on both the synthetic SUNCG-RGBD and the real NYUv2 dataset on the metric of semantic scene completion). We next detail the RAB block. 

As shown in Figure~\ref{RAB_visualization}, our RAB block consists of two sequential blocks: a DDR block and an attention block. Moreover, a shortcut connection is used to achieve the effect of the residual block~\cite{He2015DeepRL}. Our RAB can be functionally formulated as:
\begin{equation}
    y=A(D(x))+x
\end{equation}
where x is input, y is output, and D and A denote the DDR block and attention block, respectively. 
Instead of the 3D-ResNet blocks, we utilize the DDR block proposed in \cite{Li2019RGBDBD}, as shown in Figure~\ref{RAB_visualization} (b), to reduce the model parameters. The DDR block decomposes a 3D convolution with a kernel size of $3 \times 3 \times 3$ into three consecutive layers with kernel sizes of $1 \times 1 \times 3$, $1 \times 3 \times 1$ and $3 \times 1 \times 1$ respectively, so that the parameters could be reduced significantly.

The attention block used in RAB is a 3D adaption of the 2D attention block originally proposed in \cite{Woo2018CBAMCB}. It sequentially captures channel-wise attention and spatial-wise attention. The former takes into account the channel importance of the 3D features while the latter infers the spatial importance of the 3D features. Specifically, in our attention block, the 3D features of each channel are firstly aggregated by using average pooling and max pooling to generate two descriptors. Then the aggregated descriptors are forwarded to a multi-layer perceptron (MLP) to extract hidden features. The outputs of the MLP are element-wise added as the channel-wise attention. The generated channel-wise attention is multiplied by the 3D features to generate the channel-refined 3D features. 
In practice, the MLP has two hidden layers, and we choose 1/8 of the number of input feature channels as the hidden layer size. The spatial-wise attention is achieved via the following steps: the channel-refined 3D features are first aggregated by channel-wise average pooling and max pooling, and then the generated descriptors are forwarded to two convolution layers (the kernel sizes are $5 \times 5 \times 5$ and $1 \times 1 \times 1$, respectively) to generate the spatial-wise attention. In the last stage of RAB, the input 3D features are multiplied by the attention weight to achieve the 3D features with both channel-wise and spatial-wise attention injected.

\subsection{Implementation Details}
\label{details}
We follow \cite{Song2016SemanticSC} to process the data and compute the 2D-3D projection mapping for each sample in advance following SATNet~\cite{Liu2018SeeAT}. The training procedure consists of two steps. We first pre-train the 2D segmentation network with the supervision of 2D semantic segmentation ground truth, and then train the whole model end-to-end. We use cross-entropy loss and an SGD optimizer with a momentum of 0.9, a weight decay of 5e-4, and a batch size of 1. The learning rate of the 2D segmentation network and 3D scene completion network is 0.001 and 0.01, respectively. 

%=====================Figure=====================
\begin{figure*}[t]
\centering
\includegraphics[width=0.99\textwidth]{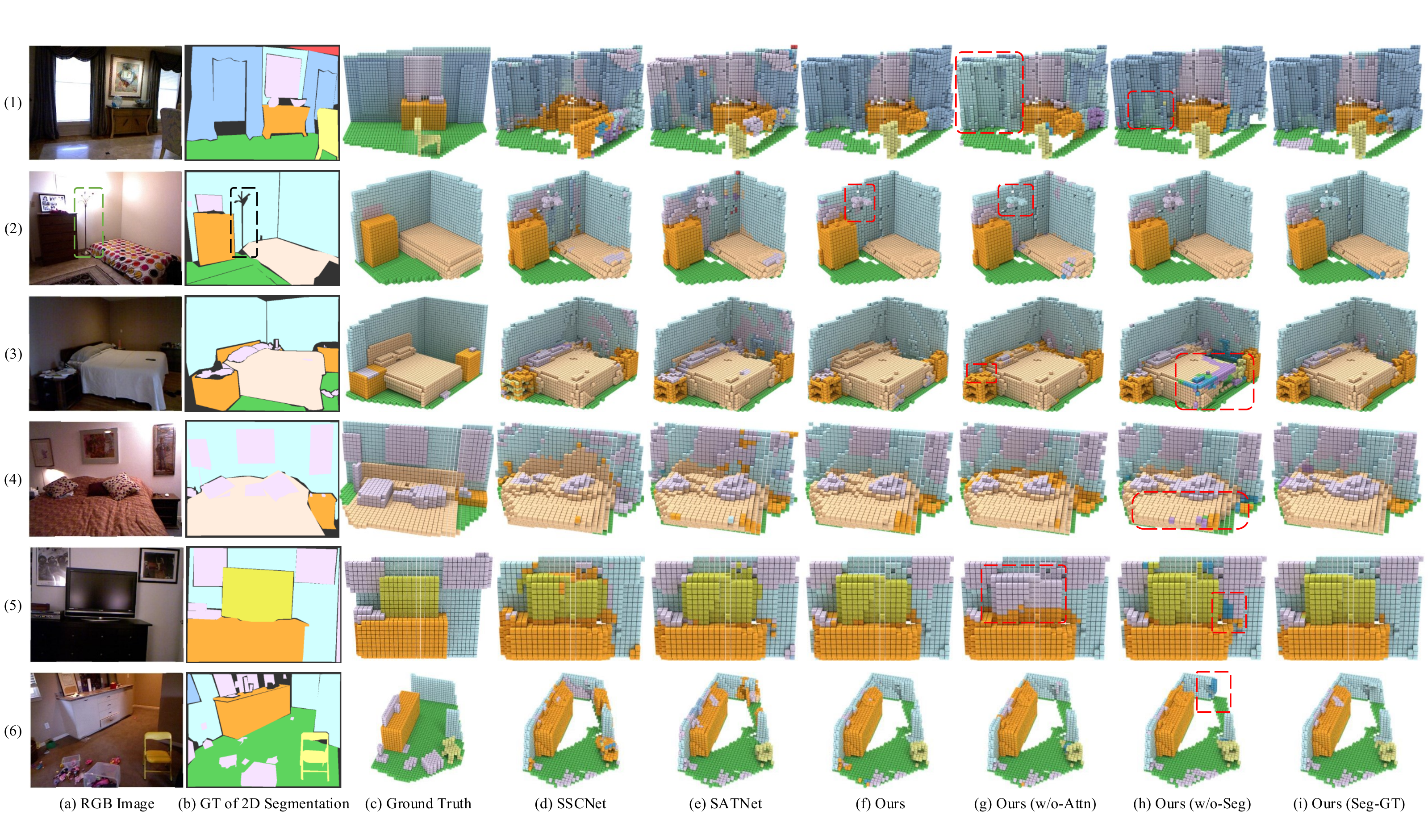}
\caption{Qualitative results on the NYUv2 dataset. From left to right: RGB image, ground truth of 2D segmentation result, ground truth of SSC task, results generated by SSCNet~\cite{Song2016SemanticSC}, SATNet~\cite{Liu2018SeeAT}, our approach, our approach with attention block removed, our approach with 3D-guidance branch removed, and our approach
with the result of the 2D semantic segmentation module (see Figure~\ref{network_structure}) replaced by the ground truth.}
\label{visualization}
\end{figure*}
%=====================Figure=====================

\section{Experiments}
In this section, the proposed method is evaluated and compared with the state-of-the-art methods on two public datasets, i.e., the NYUv2 dataset~\cite{Silberman2012IndoorSA} and the SUNCG-RGBD dataset~\cite{Liu2018SeeAT}.

\subsection{Datasets and Metrics}
\subsubsection{Datasets}
We evaluate our method on the NYUv2 and the SUNCG-RGBD datasets. The NYUv2~\cite{Silberman2012IndoorSA} is a real scene dataset, consisting of 1449 indoor scenes. The dataset is divided into 795 training and 654 testing samples, each scene associated with RGB-D images. We obtain the ground truth of the SSC task by following~\cite{Song2016SemanticSC}. NYUv2 is a challenging dataset due to the complexity of the indoor scene and the measurement errors in the depth images caused by the Kinect data collection. SUNCG-RGBD, a synthetic dataset proposed by Liu et al.~\cite{Liu2018SeeAT}, is a subset of the SUNCG dataset~\cite{Song2016SemanticSC}. It consists of 13011 training samples and 499 testing samples. 
\subsubsection{Metrics}
We mainly validate the proposed framework on two tasks, scene completion and semantic scene completion, as previous methods. We use the voxel-level intersection over union (IoU) between ground truth labels and predicted labels as the evaluation metric on both tasks. Specifically, for the task of semantic scene completion, we evaluate the IoU of each category on both the observed and occluded voxels. For the task of scene completion, we treat all non-empty voxels as the same category and evaluate the IoU of the binary predictions on the occluded voxels.

\subsection{Quantitative Comparison}
\subsubsection{Comparison on NYUv2 dataset}
Table \ref{tab1} presents the comparison results of both the scene completion and the semantic scene completion on the NYUv2 dataset. Compared with previous methods, our model achieves state-of-the-art performance on the task of semantic scene completion and ranks second on the task of scene completion. Compared to DDRNet~\cite{Li2019RGBDBD}, which also uses RGB-D images as input, our method achieves gains of $2.6\%$ on the task of semantic scene completion. The quantitative result also reveals that the proposed method demonstrates superior performance against previous methods on the categories with small physical size such as \emph{chair} (a gain of $4.9\%$ is achieved) and the categories with variational depth information such as \emph{window} (a gain of $5.1\%$ is achieved). We attribute this improvement to the application of the attention block and the guidance branch in the 3D network. The attention block can make the model focus on significant parts. Therefore, the features of small objects can be acquired more effectively and further achieve better semantic scene completion performance. Meanwhile, the guidance branch, which is boosted with the sophisticated 2D segmentation network, can improve the task of 3D semantic scene completion by providing reliable guidance information for the 3D semantic scene completion branch. In the discussion section, we will validate this explanation by ablation study.

\subsubsection{Comparison on SUNCG-RGBD dataset} 
Table~\ref{tab2} presents the performance of our method on the SUNCG-RGBD dataset with comparison to some previous approaches. 
Compared with these baseline methods, our model achieves the best performance on both the tasks of scene completion and semantic scene completion. Specifically, our method outperforms the previous SSCNet by significant performance margin, which are $9.6\%$ gains on the semantic scene completion task and $13.0\%$ gains on the scene completion task. Compared to another recent model SATNet, our method still has a distinct advantage ($2.5\%$ gains on semantic scene completion and $0.6\%$ gains on scene completion). 
Consistent with the observations on the NYUv2 dataset, our model performs well on those categories like \emph{chair} and \emph{window} and respectively achieves gains of $0.9\%$ and $4\%$ against SATNet. 

\subsection{Qualitative Analysis}
Figure~\ref{visualization} visualizes the qualitative results of the semantic scene completion task generated by the proposed method and two previous methods, SSCNet and SATNet, on a set of representative samples from the NYUv2 dataset.
It can be easily seen that our method has achieved better performance than SSCNet and SATNet. For example, our method produced much better results for objects with relatively large physical size, such as the wall and floor, in terms of segmentation consistency, as shown in row (1) and row (4). Similarly, our method can achieve a clearer 3D shape boundary of small objects, such as \emph{chair} in row (1) and \emph{obj.} in row (3). We attribute this to the guidance branch, which can provide the boundary constraints. In row (2), it can be seen from the RGB image of the scene that there is a lamp within the green dashed-line box, but it is regarded as background in the ground truth of both 2D segmentation and 3D semantic scene completion. Our model can reconstruct it well as shown in column (f) (see the red dashed-line box). This indicates that our model is capable of recognizing and reconstructing objects with small physical size in the scene.

%=====================Figure=====================
\begin{figure*}[t]
\centering
\includegraphics[width=0.99\textwidth]{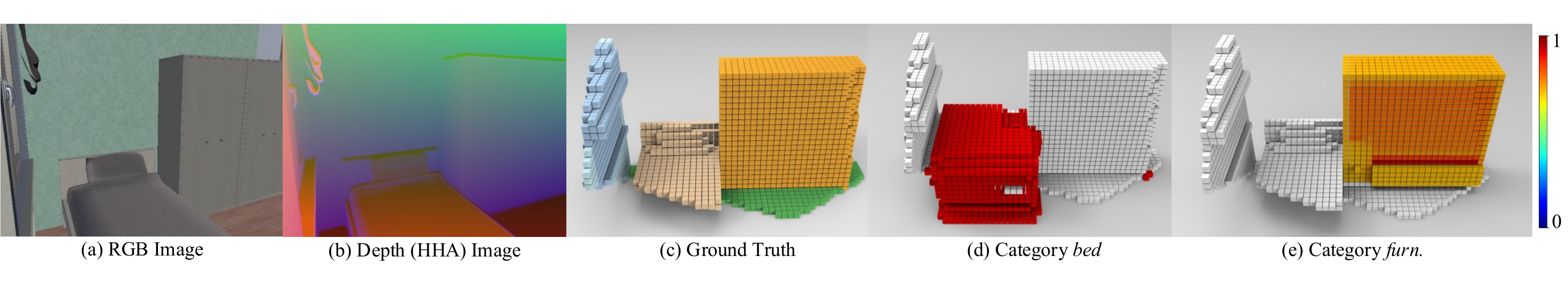}
\caption{\textbf{Visualization of the output of 3D-guidance branch.} (a-b) The RGB-D images. (c) Ground truth of SSC. (d-e) 3D segmentation results of category \emph{bed} and \emph{furn.} (visualized from the output of the 3D-guidance branch). The color bar, which quantifies the category probability, is shown on the right as a reference. It can be seen that the output of the 3D-guidance branch does provide reliable guidance information for the task of SSC.}
\label{segbranch}
\end{figure*}
%=====================Figure=====================

\section{Discussion}

\subsection{How useful is the 3D-guidance branch?}
\label{IsSegUseful}
To investigate if the 3D-guidance branch can be helpful for the semantic scene completion task, we study the performance of our method without the 3D-guidance branch. 
Specifically, we remove the 3D-guidance branch in Figure~\ref{network_structure} and leave only the 3D-semantic completion branch.
In Table \ref{tab1} and Table \ref{tab2}, we show the quantitative performance of our model without the 3D-guidance branch (denoted as Ours (w/o-Seg)).
Its performance on average for 3D semantic scene completion is 45.5\% on the SUNCG-RGBD dataset, and 29.9\% on the NYUv2 dataset. In contrast, the model with the guidance branch can achieve 1.8\% and 3.1\% performance improvements against the one with the 3D-guidance branch removed on the two datasets, respectively. Compared to the basic model (denoted as Ours (basic)) where both the attention block and the guidance branch are removed from the proposed architecture, the 3D-guidance branch can respectively improve the performance by 2.1\% and 0.7\% on the SUNCG-RGBD dataset and the NYUv2 dataset.

In Figure~\ref{segbranch}, we visualize the results of the outputs of the 3D-guidance branch. We can see that the 3D-guidance branch can accurately focus on the corresponding 3D regions and provide reliable constraints of object boundaries, which makes the 3D semantic scene completion network able to determine the accurate voxel occupancy. Therefore it benefits both the scene completion and semantic scene completion task.
In Figure~\ref{visualization}, we visually demonstrate the effect of the 3D-guidance branch. In row (3) and row (5), the model with 3D-guidance branch removed is not able to maintain the integrity of the \emph{bed} and \emph{wall}, as shown in column (h). However, with the 3D-guidance branch added, the proposed method can address this problem, as shown in column (f).

\subsection{How useful are the residual attention blocks?}
Previous works have proved that the attention mechanism can greatly help the 2D image classification task. We here investigate if the attention mechanism can benefit the 3D semantic scene completion task. To achieve this purpose, we evaluate the performance of the proposed architecture with the attention block removed (i.e., only the DDR block in Figure~\ref{RAB_visualization} (a) remains). In Table \ref{tab1} and Table \ref{tab2}, we report the performance of our model with attention mechanism removed (denoted as Ours (w/o-Attn)). The average performance of this model is 44.3\% on the SUNCG-RGBD dataset and 30.1\% on the NYUv2 dataset. Conversely, the model with the attention mechanism added can achieve 3\% and 2.9\% performance improvements on the two datasets, respectively. Compared with the basic model (denoted as Ours (basic)), the attention block can improve the performance 0.5\% and 3.3\% on the two datasets, respectively.

We guess this performance improvement is probably because the proposed RAB calculates both spatial-wise and channel-wise weights and is able to make the network focus on meaningful parts instead of empty voxels. 
Moreover, the attention mechanism in this work can improve the performance of the network on small objects, such as chairs and tables. For example, in row (3) of Figure~\ref{visualization}, it can be seen from the RGB image that there is a pile of paper on the cabinet. 
The proposed method can recognize and reconstruct it while the one with attention block removed cannot complete it completely.

\subsection{Is it possible to improve AMFNet?}

Our proposed AMFNet improves the performance of SSC by mainly leveraging 2D semantic segmentation results to provide guidance information and applying attention mechanisms to enhance the capability of feature extraction. Here we envision potential improvements in two aspects. 

(1) In Table \ref{tab1} and Table \ref{tab2}, we report the performance of the architecture which uses the ground truth of the 2D semantic segmentation as input to the 3D-guidance branch during the inferring stage (denoted as Ours (Seg-GT)). It can be seen that the ground truth segmentation can improve the performance of the proposed model by 1.8\% and 1.2\% on the SUNCG-RGBD and NYUv2 datasets respectively, which indicates an accessible way to improve our model is to enhance the performance of the 2D segmentation network. 

(2) In the 3D-guidance branch we perform a one-hot encoding to the semantic segmentation results of the visible voxels in a 3D scene, which may lead to inaccurate object constraints for the scenes where the objects belonging to the same category are spatially separated. In these scene samples, a large range of empty voxels between these objects might be encoded into the corresponding category and lead to inaccurate object boundaries. Therefore, another accessible way to improve our model is to use instance segmentation information for one-hot encoding.

\section{Conclusion}
In this paper, we introduce a novel network named AMFNet for the task of semantic scene completion. With the guidance of 2D semantic segmentation and the help of the attention focusing on reliable depth cues in the spatial dimension, AMFNet is capable of improving the completion and segmentation accuracy simultaneously by making full use of the information of the input RGB-D images. The major technique contributions of AMFNet include a two-branch fusion network boosted with 2D semantic segmentation and 3D semantic completion network empowered by residual attention blocks. Experimental results on the SUNCG-RGBD dataset and NYUv2 dataset demonstrate AMFNet achieves state-of-the-art performance. Ablation study and visualization results also show that the two-branch network and the attention-injected network have significant contributions to the proposed method.

\section{Acknowledgements}
This work was supported by the National Natural Science Funds of China (61671267).

\bibliography{AAAI-LiS.6779}

\begin{thebibliography}{}

\bibitem[\protect\citeauthoryear{Chen \bgroup et al\mbox.\egroup
  }{2016}]{Chen2016DeepLabSI}
Chen, L.-C.; Papandreou, G.; Kokkinos, I.; Murphy, K.; and Yuille, A.~L.
\newblock 2016.
\newblock {DeepLab: Semantic Image Segmentation with Deep Convolutional Nets,
  Atrous Convolution, and Fully Connected CRFs}.
\newblock {\em IEEE Transactions on Pattern Analysis and Machine Intelligence}
  40:834--848.

\bibitem[\protect\citeauthoryear{Chen \bgroup et al\mbox.\egroup
  }{2018}]{Chen2018EncoderDecoderWA}
Chen, L.-C.; Zhu, Y.; Papandreou, G.; Schroff, F.; and Adam, H.
\newblock 2018.
\newblock {Encoder-Decoder with Atrous Separable Convolution for Semantic Image
  Segmentation}.
\newblock In {\em Proceedings of the European Conference on Computer Vision},
  801--818.

\bibitem[\protect\citeauthoryear{Firman \bgroup et al\mbox.\egroup
  }{2016}]{Firman2016StructuredPO}
Firman, M.; Mac~Aodha, O.; Julier, S.; and Brostow, G.~J.
\newblock 2016.
\newblock {Structured Prediction of Unobserved Voxels from a Single Depth
  Image}.
\newblock In {\em Proceedings of the IEEE Conference on Computer Vision and
  Pattern Recognition},  5431--5440.

\bibitem[\protect\citeauthoryear{Geiger and Wang}{2015}]{Geiger2015Joint3O}
Geiger, A., and Wang, C.
\newblock 2015.
\newblock {Joint 3D Object and Layout Inference from a Single RGB-D Image}.
\newblock In {\em German Conference on Pattern Recognition},  183--195.

\bibitem[\protect\citeauthoryear{Graham, Engelcke, and van~der
  Maaten}{2018}]{Graham20173DSS}
Graham, B.; Engelcke, M.; and van~der Maaten, L.
\newblock 2018.
\newblock {3D Semantic Segmentation with Submanifold Sparse Convolutional
  Networks}.
\newblock In {\em Proceedings of the IEEE Conference on Computer Vision and
  Pattern Recognition},  9224--9232.

\bibitem[\protect\citeauthoryear{Guedes, de Campos, and
  Hilton}{2018}]{Hilton2018}
Guedes, A. B.~S.; de~Campos, T.~E.; and Hilton, A.
\newblock 2018.
\newblock {Semantic Scene Completion Combining Colour and Depth: Preliminary
  Experiments}.
\newblock {\em arXiv preprint arXiv:1802.04735}.

\bibitem[\protect\citeauthoryear{Guo and Tong}{2018}]{Tong2018}
Guo, Y.-X., and Tong, X.
\newblock 2018.
\newblock {View-volume Network for Semantic Scene Completion from a Single
  Depth Image}.
\newblock {\em arXiv preprint arXiv:1806.05361}.

\bibitem[\protect\citeauthoryear{Gupta, Arbelaez, and
  Malik}{2013}]{Gupta2013PerceptualOA}
Gupta, S.; Arbelaez, P.; and Malik, J.
\newblock 2013.
\newblock {Perceptual Organization and Recognition of Indoor Scenes from RGB-D
  Images}.
\newblock In {\em Proceedings of the IEEE Conference on Computer Vision and
  Pattern Recognition},  564--571.

\bibitem[\protect\citeauthoryear{Gupta \bgroup et al\mbox.\egroup
  }{2014}]{Gupta2014LearningRF}
Gupta, S.; Girshick, R.; Arbel{\'a}ez, P.; and Malik, J.
\newblock 2014.
\newblock {Learning Rich Features from RGB-D Images for Object Detection and
  Segmentation}.
\newblock In {\em Proceedings of the European Conference on Computer Vision},
  345--360.

\bibitem[\protect\citeauthoryear{Han \bgroup et al\mbox.\egroup
  }{2019}]{han2019deep}
Han, X.; Zhang, Z.; Du, D.; Yang, M.; Yu, J.; Pan, P.; Yang, X.; Liu, L.;
  Xiong, Z.; and Cui, S.
\newblock 2019.
\newblock {Deep Reinforcement Learning of Volume-guided Progressive View
  Inpainting for 3D Point Scene Completion from a Single Depth Image}.
\newblock In {\em Proceedings of the IEEE Conference on Computer Vision and
  Pattern Recognition},  234--243.

\bibitem[\protect\citeauthoryear{Hays and Efros}{2007}]{Hays2008SceneCU}
Hays, J., and Efros, A.~A.
\newblock 2007.
\newblock {Scene Completion Using Millions of Photographs}.
\newblock {\em ACM Transactions on Graphics} 26(3):4.

\bibitem[\protect\citeauthoryear{He \bgroup et al\mbox.\egroup
  }{2016}]{He2015DeepRL}
He, K.; Zhang, X.; Ren, S.; and Sun, J.
\newblock 2016.
\newblock {Deep Residual Learning for Image Recognition}.
\newblock In {\em Proceedings of the IEEE Conference on Computer Vision and
  Pattern Recognition},  770--778.

\bibitem[\protect\citeauthoryear{Hu, Shen, and
  Sun}{2018}]{Hu2017SqueezeandExcitationN}
Hu, J.; Shen, L.; and Sun, G.
\newblock 2018.
\newblock {Squeeze-and-Excitation Networks}.
\newblock In {\em Proceedings of the IEEE Conference on Computer Vision and
  Pattern Recognition},  7132--7141.

\bibitem[\protect\citeauthoryear{Laugier \bgroup et al\mbox.\egroup
  }{2011}]{LaugierPPYYTMN11}
Laugier, C.; Paromtchik, I.~E.; Perrollaz, M.; Yong, M.; Yoder, J.-D.; Tay, C.;
  Mekhnacha, K.; and N{\`e}gre, A.
\newblock 2011.
\newblock {Probabilistic Analysis of Dynamic Scenes and Collision Risks
  Assessment to Improve Driving Safety}.
\newblock {\em IEEE Intelligent Transportation Systems Magazine} 3(4):4--19.

\bibitem[\protect\citeauthoryear{Li \bgroup et al\mbox.\egroup
  }{2019a}]{li2019dfanet}
Li, H.; Xiong, P.; Fan, H.; and Sun, J.
\newblock 2019a.
\newblock {Dfanet: Deep Feature Aggregation for Real-time Semantic
  Segmentation}.
\newblock In {\em Proceedings of the IEEE Conference on Computer Vision and
  Pattern Recognition},  9522--9531.

\bibitem[\protect\citeauthoryear{Li \bgroup et al\mbox.\egroup
  }{2019b}]{Li2019RGBDBD}
Li, J.; Liu, Y.; Gong, D.; Shi, Q.; Yuan, X.; Zhao, C.; and Reid, I.
\newblock 2019b.
\newblock {RGBD Based Dimensional Decomposition Residual Network for 3D
  Semantic Scene Completion}.
\newblock In {\em Proceedings of the IEEE Conference on Computer Vision and
  Pattern Recognition},  7693--7702.

\bibitem[\protect\citeauthoryear{Lin, Fidler, and
  Urtasun}{2013}]{Lin2013HolisticSU}
Lin, D.; Fidler, S.; and Urtasun, R.
\newblock 2013.
\newblock {Holistic Scene Understanding for 3D Object Detection with RGBD
  Cameras}.
\newblock In {\em Proceedings of the IEEE International Conference on Computer
  Vision},  1417--1424.

\bibitem[\protect\citeauthoryear{Liu \bgroup et al\mbox.\egroup
  }{2018}]{Liu2018SeeAT}
Liu, S.; Hu, Y.; Zeng, Y.; Tang, Q.; Jin, B.; Han, Y.; and Li, X.
\newblock 2018.
\newblock {See and Think: Disentangling Semantic Scene Completion}.
\newblock In {\em Advances in Neural Information Processing Systems},
  263--274.

\bibitem[\protect\citeauthoryear{Ren, Bo, and Fox}{2012}]{Ren2012RGBDSL}
Ren, X.; Bo, L.; and Fox, D.
\newblock 2012.
\newblock {RGB-(D) Scene Labeling: Features and Algorithms}.
\newblock In {\em Proceedings of the IEEE Conference on Computer Vision and
  Pattern Recognition},  2759--2766.

\bibitem[\protect\citeauthoryear{Silberman \bgroup et al\mbox.\egroup
  }{2012}]{Silberman2012IndoorSA}
Silberman, N.; Hoiem, D.; Kohli, P.; and Fergus, R.
\newblock 2012.
\newblock {Indoor Segmentation and Support Inference from RGBD Images}.
\newblock In {\em Proceedings of the European Conference on Computer Vision},
  746--760.

\bibitem[\protect\citeauthoryear{Song \bgroup et al\mbox.\egroup
  }{2017}]{Song2016SemanticSC}
Song, S.; Yu, F.; Zeng, A.; Chang, A.~X.; Savva, M.; and Funkhouser, T.
\newblock 2017.
\newblock {Semantic Scene Completion from a Single Depth Image}.
\newblock In {\em Proceedings of the IEEE Conference on Computer Vision and
  Pattern Recognition},  1746--1754.

\bibitem[\protect\citeauthoryear{Wang \bgroup et al\mbox.\egroup
  }{2017}]{Wang2017ResidualAN}
Wang, F.; Jiang, M.; Qian, C.; Yang, S.; Li, C.; Zhang, H.; Wang, X.; and Tang,
  X.
\newblock 2017.
\newblock {Residual Attention Network for Image Classification}.
\newblock In {\em Proceedings of the IEEE Conference on Computer Vision and
  Pattern Recognition},  3156--3164.

\bibitem[\protect\citeauthoryear{Woo \bgroup et al\mbox.\egroup
  }{2018}]{Woo2018CBAMCB}
Woo, S.; Park, J.; Lee, J.-Y.; and So~Kweon, I.
\newblock 2018.
\newblock {CBAM: Convolutional Block Attention Module}.
\newblock In {\em Proceedings of the European Conference on Computer Vision},
  3--19.

\bibitem[\protect\citeauthoryear{Xiong \bgroup et al\mbox.\egroup
  }{2019}]{xiong2019upsnet}
Xiong, Y.; Liao, R.; Zhao, H.; Hu, R.; Bai, M.; Yumer, E.; and Urtasun, R.
\newblock 2019.
\newblock {Upsnet: A Unified Panoptic Segmentation Network}.
\newblock In {\em Proceedings of the IEEE Conference on Computer Vision and
  Pattern Recognition},  8818--8826.

\bibitem[\protect\citeauthoryear{Yang \bgroup et al\mbox.\egroup
  }{2018}]{yang2018denseaspp}
Yang, M.; Yu, K.; Zhang, C.; Li, Z.; and Yang, K.
\newblock 2018.
\newblock {Denseaspp for Semantic Segmentation in Street Scenes}.
\newblock In {\em Proceedings of the IEEE Conference on Computer Vision and
  Pattern Recognition},  3684--3692.

\bibitem[\protect\citeauthoryear{Zhang \bgroup et al\mbox.\egroup
  }{2018}]{Zhang2018EfficientSS}
Zhang, J.; Zhao, H.; Yao, A.; Chen, Y.; Zhang, L.; and Liao, H.
\newblock 2018.
\newblock {Efficient Semantic Scene Completion Network with Spatial Group
  Convolution}.
\newblock In {\em Proceedings of the European Conference on Computer Vision},
  733--749.

\end{thebibliography}
\bibliographystyle{aaai}

\end{document}